\documentclass[sigconf, screen, nonacm]{acmart}

\AtBeginDocument{%
  }

\usepackage{tabularx}
\usepackage{multirow}
\usepackage{color,soul}
\usepackage{comment}
\usepackage{booktabs}
\usepackage{array}
\usepackage{xcolor}

\definecolor{promptorange}{HTML}{E67E22} 
\definecolor{jsonblue}{HTML}{2980B9}     
\definecolor{jsongray}{HTML}{7F8C8D}     

\definecolor{highlightorange}{HTML}{E67E22} 
\definecolor{highlightblue}{HTML}{2980B9}   

\definecolor{badgegreen}{HTML}{27AE60}
\definecolor{badgered}{HTML}{C0392B}

\newcolumntype{Y}{>{\raggedright\arraybackslash}X}
\newcolumntype{Z}{>{\raggedright\arraybackslash}X}
\begin{document}

\title{From Prompting to Behavioral Alignment: Personalized LLM Judges for Recommendation Evaluation}

\author{Alireza S. Ziabari}
\affiliation{%
  \institution{Netflix}
  \city{Los Angeles}
  \country{USA}
}
\email{asalkhordehziabari@netflix.com}

\author{Kat Ellis}
\affiliation{%
  \institution{Netflix}
  \city{Los Gatos}
  \country{USA}}
\email{kkatherineellis@netflix.com}

\author{Colleen Chan}
\affiliation{%
  \institution{Netflix}
  \city{Los Gatos}
  \country{USA}}
\email{colleenc@netflix.com}

\author{Ding Tong}
\affiliation{%
  \institution{Netflix}
  \city{Los Gatos}
  \country{USA}}
\email{dingt@netflix.com}

\renewcommand{\shortauthors}{Ziabari et al.}


\begin{abstract}
Traditional offline recommendation evaluation relies heavily on complex, manually maintained feature pipelines that are difficult to scale. While Large Language Models (LLMs) offer a promising alternative by predicting user engagement directly from raw text logs, empirical analysis in this study identifies a critical failure mode termed bidirectional rationalization. In a zero-shot setting, LLMs are found to convincingly argue for both positive and negative user engagement outcomes on the exact same item with identical evidence, highlighting the unreliability of off-the-shelf LLMs in predicting user engagement. To resolve this, we develop and apply a sequential behavioral alignment framework pairing fine-tuning with preference optimization over paired correct and counterfactual rationales. Evaluated on real-world homepage interaction logs, this aligned reasoning approach achieves a 32.19\% lift in Macro-F1 score over the zero-shot baseline and matches the production feature-engineered baseline. The results demonstrate that behavioral alignment mitigates bidirectional rationalization while delivering human-interpretable reasoning traces without manual pipeline overhead.
\end{abstract}



\maketitle

\section{Introduction}\label{sec:intro}

Recommendation systems are a core component of modern digital platforms, driving user engagement and long-term retention \cite{gomez2015netflix}. To continuously improve these systems, rapid and reliable evaluation frameworks are essential. While online A/B testing remains the gold standard for measuring true user satisfaction and business impact, it is inherently slow and computationally intensive and risks exposing users to suboptimal experiences. Consequently, industrial practitioners rely heavily on offline evaluation using historical interaction logs. However, traditional offline metrics (such as NDCG or Recall) often suffer from exposure bias and frequently fail to correlate strongly with online A/B test results \cite{gilotte2018offline}. This persistent gap between offline and online performance creates a significant bottleneck for rapid model iteration.

To bridge this evaluation gap, the industry has increasingly adopted model-based evaluation: deploying learned models that estimate counterfactual outcomes and business metrics \cite{zangerle2022evaluating}. Furthermore, recent work has explored using LLMs as offline judges \cite{zheng2023judging}. Unlike standard retrieval metrics, LLMs possess the semantic capacity to interpret complex contextual signals. Building on the promise of LLM-based evaluation, our work focuses on enhancing the rigor and discriminative capacity of these models. However, deploying LLMs for recommendation evaluation introduces the critical challenge of personalization. Traditional LLM judges assess general qualities, such as helpfulness or safety, based on global human consensus. In contrast, recommendation quality is inherently subjective -- an item's relevance depends on individual user preferences rather than a universal standard. Consequently, generic evaluation criteria fail to capture true user intent, making scalable, personalized evaluation a primary bottleneck for industrial applications.

Within the context of the Netflix recommendation system, we frame this personalized evaluation as an engagement prediction task. Specifically, when a user is presented with a curated row of recommended titles on their homepage, the evaluator must analyze the user's chronological viewing history and immediate session context to predict a binary behavioral outcome: whether the user will engage with the recommendation (a ``play'' action) or ignore the row entirely (a ``skip'' action). While out-of-the-box LLMs possess the semantic capability to read this serialized history, they inherently struggle to act as reliable judges. Without domain-specific alignment, these models exhibit biased evaluations \cite{wang2024large}. Furthermore, unaligned models often act as unconstrained rationalizers \cite{sharma2023towards}; therefore, when evaluating generally high-quality recommendation rows, an unaligned LLM can effortlessly construct plausible justifications for either a ``play'' or a ``skip'' action, regardless of the true user intent. 

To address this, we propose an end-to-end framework that aligns an LLM judge's reasoning to grounded user engagement. Rather than relying on zero-shot prompting, we train the model to generate Chain-of-Thought (CoT) \cite{wei2022chain} reasoning traces anchored to actual user engagement outcomes, and systematically benchmark inference-time and parameter-level adaptation strategies for this task.

In this paper, we characterize a personalization-specific failure mode of LLM-based evaluators, and show that \emph{behavioral alignment} using preference optimization that anchors the LLM judge's reasoning to observed user engagement can match a feature-engineered production baseline while preserving interpretability. Our main contributions are as follows:

\begin{itemize}
\item We identify \emph{bidirectional rationalization} as a personalization-specific failure mode of LLM judges that is structurally distinct from hallucination. This failure traces to foundational recommender-system trade-offs (e.g., short-term vs.\ long-term, accuracy vs.\ diversity, novelty vs.\ popularity, exploration vs.\ exploitation) each of which admits multiple defensible reasoning pathways that an unaligned model can elaborate into a fluent argument in either direction. After filtering rationale pairs for unfactual claims, the bidirectional disagreement patterns persist, showing that the failure is not reducible to fabrication.

\item Through a systematic comparison of prompting strategies, we find that reasoning-based prediction and the inclusion of immediate session context are the only consistent contributors to LLM judge accuracy. However, prompt engineering alone is insufficient to close the gap between a zero-shot LLM judge and a heavily feature-engineered baseline, motivating the move to parameter-level adaptation.

\item We propose an alignment recipe that first applies SFT on reasoning traces, and then applies offline preference optimization over paired correct and counterfactual reasonings grounded in true engagement outcomes. This recipe closes the remaining performance gap: the resulting text-based LLM evaluator matches the feature-engineered baseline on Netflix homepage engagement prediction without any manual feature engineering, while producing human-interpretable reasoning traces that reveal the user-history signals driving each prediction.
\end{itemize}

\section{Related Work}\label{sec:relatedwork}

\paragraph{LLM-as-judge for recommendation evaluation.}
LLMs have been used as offline evaluators across a broad range of tasks, typically by issuing direct or pairwise judgments over candidate outputs \cite{zheng2023judging}. In the recommendation setting specifically, the Profile-Aware LLM judge \cite{fabbri2025evaluating} shows that prompting LLMs with user profiles can yield judgments that approximate human ratings. A complementary line evaluates LLM-based conversational recommender systems by measuring how closely the system's recommendation strategies agree with those of human recommenders \cite{yang2024behavior}. Our work extends these lines along two axes: the judgment target shifts from a human-annotated label to an observed behavioral outcome (whether the user engages with the recommended row), and the adaptation method moves from inference-time prompting to behavioral alignment via preference optimization over reasoning rationales.

\paragraph{Reasoning for recommendation tasks.}
A parallel line of work integrates explicit reasoning into recommendation models, sharing our goal of improving both accuracy and interpretability. OneRec-Think~\cite{liu2025onerec} introduces chain-of-thought reasoning into generative recommendation, producing human-interpretable rationales alongside item predictions and demonstrating gains in live deployment. Related approaches such as ThinkRec \cite{yu2026thinkrec} and Reason-to-Recommend \cite{zhao2025reason} use supervised fine-tuning and reinforcement learning to instill reasoning capabilities in generative recommenders, with the explicit reasoning traces serving as a transparency mechanism over the model's decision process. Our work applies reasoning to the complementary problem of \emph{judging} (offline evaluation) rather than generating recommendations: the reasoning trace produced by our judge is human-readable and exposes which user-history signals drove each predicted engagement outcome, providing an interpretable alternative to opaque feature-based evaluators.

\paragraph{Bidirectional rationalization in LLM judges.}
Unaligned LLMs have been characterized as unconstrained rationalizers that can elaborate plausible justifications in either direction on ambiguous inputs \cite{sharma2023towards}. Recent work shows that two reasoning models given opposing positions on the same topic each produce confident, internally coherent arguments \cite{prasad2025debate}, and that chain-of-thought traces often do not faithfully reflect the actual decision process even on objective tasks with stable ground truth \cite{turpin2023language, chen2025reasoning}. We build on this literature in Section \ref{sec:rationalizer}, arguing that the rationalizer failure mode takes a structurally different form in personalized recommendation evaluation than in the objective settings studied to date.

\begin{table*}[t]
\caption{Same user history, same recommended row, two reasoning paths reaching opposite conclusions. After manually filtering rationale pairs for unfactual claims, these bidirectional disagreement patterns persist and trace to foundational recommender-system trade-offs. Each path's predicted output and the true engagement label are annotated; reasoning excerpts are abridged from teacher-generated rationales.}
\label{tab:rationalizer-examples}
\centering
\small
\renewcommand{\arraystretch}{1.4}
\begin{tabularx}{\textwidth}{@{}p{0.21\textwidth}YY@{}}
\toprule
\textbf{Category / Trade-off} & \textbf{Reasoning for Play} & \textbf{Reasoning for Skip} \\
\midrule

\textbf{Temporal lens}\newline
\textcolor{highlightorange}{\textbf{Short-term}} vs.\ \textcolor{highlightblue}{\textbf{long-term}}\newline\newline
User Engagement: \color{badgered}{\textbf{SKIP}}
& \textbf{Recent-anchored.} ``\dots\textit{\textbf{\color{highlightorange}{Recent interactions show a preference}} for movies and TV shows that are light-hearted and inspiring}\dots\ likely that the user will play the recommended movies, particularly those with romantic and comedic tones.''
& \textbf{Long-term.} ``Strong preference for romance and drama, high engagement with titles with a strong emotional tone\dots\ \textit{the \textbf{\color{highlightblue}{user's viewing history suggests}} they prefer more intense and emotional content}\dots\ likely to skip.'' \\

\midrule
\textbf{Multi-modal taste}\newline
\textcolor{highlightorange}{\textbf{Calibration}} vs.\ \textcolor{highlightblue}{\textbf{Specialization}}\newline\newline
User Engagement: \color{badgegreen}{\textbf{PLAY}}
& \textbf{Fantasy/drama lens.} ``\textbf{\color{highlightorange}{Interest in fantasy, drama, and romance genre}}. The user watched \textit{\textless TITLE-1\textgreater} for 96\% of its duration, indicating strong engagement.''
& \textbf{Action/thriller lens.} ``\textbf{\color{highlightblue}{Preference for action-packed and thrilling content}} (\textit{\textless TITLE-2\textgreater}, \textit{\textless TITLE-3\textgreater}); also enjoys fantasy and romance. Recommended thrillers/dramas with darker tone don't align.'' \\

\midrule
\textbf{Platform exposure}\newline
\textcolor{highlightorange}{\textbf{Popularity}} vs.\ \textcolor{highlightblue}{\textbf{Novelty}}\newline\newline
User Engagement: \color{badgered}{\textbf{SKIP}}
& \textbf{Popularity-leaning.} ``Strong preference for animated content, \textit{particularly from \textbf{\color{highlightorange}{popular franchises like \textless TITLE-4\textgreater, \textless TITLE-5\textgreater, and \textless TITLE-6\textgreater}}}\dots\ recommended kids content fits these preferences.''
& \textbf{Niche-fit.} ``User has \textit{tendency to skip movies that are more serious or dramatic}; \textit{\textless TITLE-7\textgreater}, while popular, is gentle and child-friendly---it \textbf{\color{highlightblue}{does not match the user's specific humor/adventure preferences}} within the kids' genre.'' \\

\midrule
\textbf{Interest extension}\newline
\textcolor{highlightorange}{\textbf{Exploration}} vs.\ \textcolor{highlightblue}{\textbf{Exploitation}}\newline\newline
User Engagement: \color{badgegreen}{\textbf{PLAY}}
& \textbf{Exploration framing.} ``Comedy and action fan with irreverent/deadpan tones\dots\ also watched documentary and thriller content, indicating \textit{\textbf{\color{highlightorange}{willingness to explore different genres}}}.''
& \textbf{Exploitation framing.} ``\textit{\textbf{\color{highlightblue}{Tendency to skip movies that are too long, intense, or have a dark tone}}}. Favors comedies with lighter tone.'' \\

\bottomrule
\end{tabularx}
\end{table*}

\section{Preliminaries: Bidirectional Rationalization} \label{sec:rationalizer}

To understand the bottleneck in applying LLMs to recommendation evaluation, it is critical to characterize how unaligned models fail in personalized settings. Prior work has characterized rationalization, position bias, and sycophancy in settings with stable, externally verifiable ground truth. However, recommendation quality is inherently subjective. To investigate failure modes of personalized recommendation evaluation, we conducted a qualitative analysis over model's rational for their evaluation.

To ground our analysis, we sampled real-world homepage interaction logs where users were exposed to curated recommendation rows in Netflix homepage. Ground-truth engagement was determined using a spatial scroll heuristic: if a user played an item from a specific row, that row was logged as a positive ``play'' event, while any rows positioned above it, which the user explicitly scrolled past, were logged as hard-negative ``skip'' events. Because all recommended items were generally highly relevant to the user, these ``skip'' cases act as strong counterfactuals rather than trivial mismatches. For a balanced set of these instances, we serialized the user's interaction history, immediate session context, and the recommended row, and provided them to a highly capable, unaligned LLM. By prompting the model to justify both candidate outcomes, we extracted one confident ``play'' rationale and one confident ``skip'' rationale per instance, creating a dataset where exactly one reasoning path per pair matches the true user action.

To rule out fabrication as the source of disagreement, we filtered out cases where either rationale contained mischaracterized recommended items, fabricated user-history events, or invented unsupported behavioral traits. 
Surprisingly, 77.0\% (960 of 1{,}246) of the balanced pairs survived this factuality filter. This demonstrates that the failure mode is not hallucination; rather, the model is capable of constructing internally coherent, factually grounded arguments in opposite directions from identical evidence.

Examining these persisting disagreements, we find that the bidirectional pathways correspond systematically to foundational recommender-system trade-offs:

\begin{itemize}
    \item \textit{Short-term vs.\ long-term (temporal lens).} Recent watches and long-term history are both valid signals about the user at evaluation time.
    \item \textit{Calibration vs.\ specialization (multi-modal taste).} Users typically exhibit multiple coexisting tastes, and recommended items are heterogeneous. A reasoning path can either calibrate to the user's full taste distribution \cite{steck2018calibrated} or a single facet.
    \item \textit{Novelty vs.\ popularity (platform exposure).} Engagement can be predicted by emphasizing the user's engagement with popular, well-known items or the user's preference for niche content.
    \item \textit{Exploration vs.\ exploitation (interest extension).} A user's history can support either exploration of new and serendipitous interests or exploitation of comfort-zone preferences.
\end{itemize}

Among the 960 surviving pairs from factuality filters, 95.4\% (916) exhibit contrasts that fit one or a combination of the four trade-offs above. Table~\ref{tab:rationalizer-examples} illustrates each trade-off with examples from our filtered rationale set.

The underlying mechanism is \emph{bidirectional inference}: the same observed user signal supports opposite conclusions about future engagement depending on which trade-off a reasoning path adopts. Because an unaligned LLM judge lacks behavioral alignment to real user engagement, it has no principled basis for resolving these trade-offs and acts as an unconstrained rationalizer.

Because bidirectional rationalization survives fabrication-filtering, methods that only suppress hallucinations are insufficient. The model must learn which of the many locally valid framings most reliably predicts grounded user behavior. This structural challenge motivates the two primary research questions:

\begin{itemize}
    \item \textbf{RQ1 (Inference-Time Adaptation):} To what extent can prompt engineering mitigate bidirectional rationalization and improve the accuracy of personalized engagement prediction?
    \item \textbf{RQ2 (Parameter-Level Alignment):} Which behavioral alignment paradigm (e.g., Supervised Fine-Tuning or Direct Preference Optimization) most effectively closes the performance gap and resolves the failure modes that prompt engineering cannot address?
\end{itemize}

The fact that bidirectional rationalization survives fabrication-filtering has direct implications for alignment. Supervised fine-tuning that suppresses fabricated user traits or mischaracterized items addresses the hallucination failure mode but leaves the rationalizer problem intact, because the bidirectional patterns we identify are not artifacts of invention but structural properties of the task. Behavioral alignment offers a targeted intervention: by exposing the model to paired correct and counterfactual rationales anchored to true engagement outcomes, preference optimization teaches the model \emph{which of the many locally-valid framings most reliably predicts grounded user behavior}, collapsing the bidirectional rationalizer into a directional judge that resolves the named trade-offs in the direction real users exhibit. Behavioral alignment is therefore a critical component for reliable personalized LLM judges.

\begin{figure}
    \centering
    \includegraphics[width=1\linewidth]{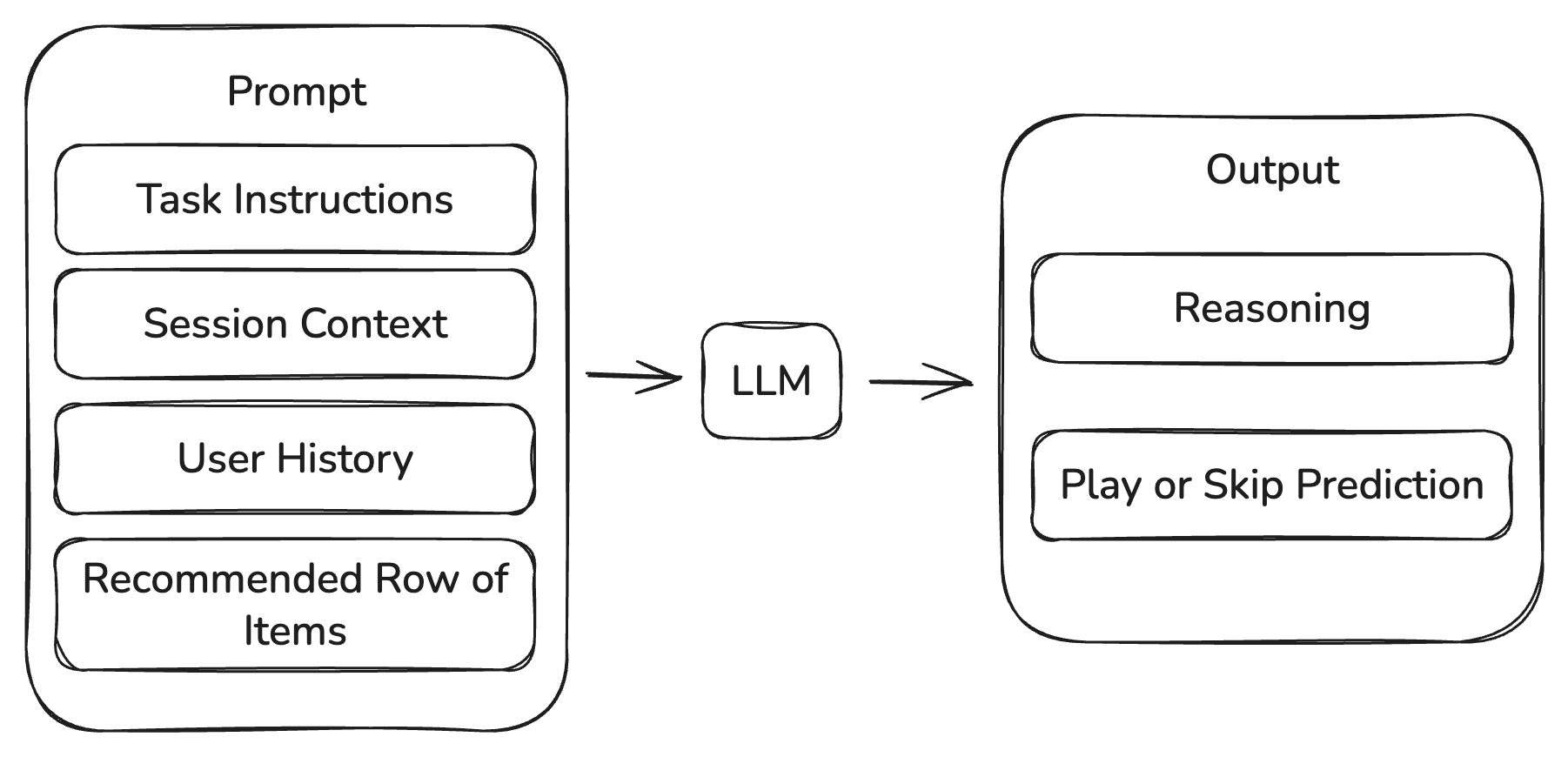}
    \caption{Task set-up for user engagement prediction. LLM receives a structured prompt containing task instructions, session context, user history, and a row of recommended items, and outputs a binary play or skip prediction and optionally, reasoning, depending on the prompt template. }
    \label{fig:task_setup}
\end{figure}

\section{Behavioral Alignment} \label{sec:method}

We detail our methodology for delivering behavioral alignment to the LLM judge. The methodology moves incrementally from zero-shot LLM usage toward progressively stronger forms of adaptation. We first describe the task set-up and initial formulation of text prompts, then outline how we generate training datasets using a combination of user interaction data and a powerful reasoning LLM, and finally present training strategies implemented to adapt LLMs for this specific task.

\subsection{Task Set-up}

We formulate our task as a binary prediction: given a recommended row of items, the user's interaction history, and relevant contextual signals, predict whether the user will engage with at least one item from that row ($\hat{y} = \text{play}$) or ignore it ($\hat{y} = \text{skip}$).

To enable the LLM to process this prediction task, we construct a structured natural language input prompt from the raw backend logs. This input prompt is composed of four main elements: the task instructions, the recommended row of items, the immediate session context (such as the current time and device), and the serialized user history. To construct the serialized user history, we convert historical interaction data such as timestamps, watch durations, and item titles into a chronological text sequence. Finally, to manage the model's context window and evaluate the impact of sequence length on predictive performance, we enforce a fixed threshold on the user history (by number of events), truncating older interactions.

\subsection{Inference and Prompting Formats} \label{sec:reasoning_prompts}

To address RQ1, a diverse suite of prompting paradigms was evaluated, ranging from straightforward zero-shot classification to highly structured, feature-aligned reasoning frameworks. These configurations are categorized into three broad tactical groups: direct label prediction (\emph{Simple} and its inverted variant \emph{Simple\textsuperscript{-1}}), basic reasoning blocks (\emph{Reason} and \emph{Reason\textsuperscript{-1}}), and specialized context-aware alignments targeting explicit behavioral signals (\emph{Evidence}, \emph{Pros and Cons}, \emph{Habit}, \emph{Persona}, \emph{Timing}, and \emph{Pattern}). The complete structural text prompts and specific JSON target schemas for all configurations are detailed in Table \ref{tab:prompt-formats}.

\begin{table*}[t]
\centering
\footnotesize
\renewcommand{\arraystretch}{1.4}
\begin{tabularx}{\textwidth}{@{}p{1cm}p{4cm}Y Z@{}}
\toprule
\textbf{Template} & \textbf{Description \& Usecase} & \textbf{Important Part of Prompt} & \textbf{Output Format} \\
\midrule

\textbf{Simple} or \textbf{Simple\textsuperscript{-1}} & 
Establishes a zero-shot classification baseline. Evaluates direct engagement prediction without reasoning overhead; toggles the target question framing to detect formatting symmetry and polarity bias. & 
\ttfamily ... Your task is to determine if the user is likely to \textcolor{promptorange}{\textbf{play (or skip)}} any of the recommended movies based on their recent activity.\newline
...\newline
Based on the user's recent activity, is the user likely to play any of the recommended movies? \textcolor{promptorange}{\textbf{Please only respond with 'Yes' or 'No'.}} & 
\ttfamily \textcolor{jsonblue}{'Yes' or 'No'} \\

\midrule
\textbf{Reason} or \textbf{Reason\textsuperscript{-1}} & 
Captures basic, unconstrained natural-language rationales. Tests if forcing an intermediate open-text justification bottleneck improves final label probability calibration. & 
\ttfamily ... Analyze user viewing patterns and \textcolor{promptorange}{\textbf{predict if they'll play (or skip)}} recommended movies. & 
\ttfamily \{\newline
\ \ \ \textcolor{jsongray}{"reasoning"}: "brief analysis",\newline
\ \ \ \textcolor{jsongray}{"result"}: \textcolor{jsonblue}{"Play or Skip"}\newline
\} \\

\midrule
\textbf{Evidence} & 
Enforces rigorous factual grounding. Instructs the judge to isolate explicit verification cues (such as duration bookmark triggers) from raw interaction logs to validate predictions. & 
\ttfamily ... Task: decide whether the user is \textcolor{promptorange}{\textbf{LIKELY (probability >= 0.80)}} to play a movie from the recommended movies based on user recent interactions.\newline
...\newline
\textbf{\#\#\# Reasoning guidelines:} ... & 
\ttfamily \{\newline
\ \ \ \textcolor{jsongray}{"evidence"}: [\{\newline
\ \ \ \ \ \ \ \textcolor{jsongray}{"signal"}: "[name of cue]",\newline
\ \ \ \ \ \ \ \textcolor{jsongray}{"event\_title"}: "[title]",\newline
\ \ \ \ \ \ \ \textcolor{jsongray}{"rec\_title"}: "[title]",\newline
\ \ \ \ \ \ \ \textcolor{jsongray}{"details"}: "[under 40 chars]"\}],\newline
\ \ \ \textcolor{jsongray}{"probability"}: "[0.00-1.00]",\newline
\ \ \ \textcolor{jsongray}{"result"}: \textcolor{jsonblue}{"[Yes or No]"} \newline
\} \\

\midrule
\textbf{Pros} \& \textbf{Cons} & 
Targets bidirectional ambiguity. Explicitly forces the model to extract, de-duplicate, and contrast competing counter-signals (skip-cues vs.\ play-cues) prior to label selection. & 
\ttfamily ... Your task is to evaluate the user's recent interactions and the recommended movies to determine if the user is \textcolor{promptorange}{\textbf{likely to play or skip}} any of the recommended movies.\newline
... & 
\ttfamily \{\newline
\ \ \ \textcolor{jsongray}{"skip\_signals"}: [less than 25 words each],\newline
\ \ \ \textcolor{jsongray}{"play\_signals"}: [less than 25 words each],\newline
\ \ \ \textcolor{jsongray}{"reasoning"}: "[less than 50 words]",\newline
\ \ \ \textcolor{jsongray}{"probability"}: "[0.00-1.00]",\newline
\ \ \ \textcolor{jsongray}{"result"}: \textcolor{jsonblue}{"[Skip or Play]"} \newline
\} \\

\midrule
\textbf{Habit} & 
Analyzes structured behavioral consistency. Directs the model to compute longitudinal trends, applying exponential recency decay weights and verifying time-of-day/day-of-week context matching. & 
\ttfamily ... Task: Determine if the user is likely to play any recommended movies by \textcolor{promptorange}{\textbf{analyzing temporal patterns}} and contextual viewing behavior.\newline
...\newline
\textbf{\#\#\# TEMPORAL ANALYSIS FRAMEWORK:}\newline
\ \ \ \ - Temporal Context Patterns\newline
\ \ \ \ - Seasonal/Trend Considerations\newline
\textbf{\#\#\# ANALYSIS STEPS} ... & 
\ttfamily \{\newline
\ \ \ \textcolor{jsongray}{"recent\_patterns"}: ["pattern1", "pattern2"],\newline
\ \ \ \textcolor{jsongray}{"temporal\_habits"}: ["habit1", "habit2"],\newline
\ \ \ \textcolor{jsongray}{"engagement\_momentum"}: "high|medium|low",\newline
\ \ \ \textcolor{jsongray}{"temporal\_match\_score"}: "0.00-1.00",\newline
\ \ \ \textcolor{jsongray}{"recency\_weighted\_probability"}: "0.00-1.00",\newline
\ \ \ \textcolor{jsongray}{"result"}: \textcolor{jsonblue}{"Play or Skip"}\newline
\} \\

\midrule
\textbf{Persona} & 
Models macro-level user archetypes. Requires the judge to map past logs to an explicit behavioral persona (e.g., Binge Watcher, Explorer) and invoke targeted evaluation heuristics. & 
\ttfamily ... Task: First \textcolor{promptorange}{\textbf{classify the user's viewing persona}}, then apply persona-specific logic to predict if they'll play any recommended movies.\newline
...\newline
\textbf{\#\#\# USER PERSONA CLASSIFICATION:}\newline
\ \ \ \ - \textbf{BINGE\_WATCHER} (Definition: ...)\newline
\ \ \ \ - \textbf{GENRE\_FOCUSED} (Definition: ...)\newline
\ \ \ \ - \textbf{CASUAL\_VIEWER} (Definition: ...) ... & 
\ttfamily \{\newline
\ \ \ \textcolor{jsongray}{"identified\_persona"}: "...",\newline
\ \ \ \textcolor{jsongray}{"persona\_confidence"}: "0.00-1.00",\newline
\ \ \ \textcolor{jsongray}{"persona\_signals"}: ["signal1", "signal2"],\newline
\ \ \ \textcolor{jsongray}{"persona\_logic"}: "reasoning for persona type",\newline
\ \ \ \textcolor{jsongray}{"recommendation\_fit"}: "0.00-1.00",\newline
\ \ \ \textcolor{jsongray}{"result"}: \textcolor{jsonblue}{"Play or Skip"}\newline
\} \\

\midrule
\textbf{Timing} & 
Isolates transient session mechanics. Anchors predictions heavily on immediate temporal variables, current interaction spikes, and rolling short-term engagement momentum. & 
\ttfamily ... Task: Predict if the user will play any recommended movies by analyzing their recent viewing behavior and timing patterns.\newline
...\newline
Focus on \textcolor{promptorange}{\textbf{recent activity (last 7 days get highest weight)}}, viewing time patterns, and engagement momentum.\newline
... & 
\ttfamily \{\newline
\ \ \ \textcolor{jsongray}{"recent\_activity"}: "high|medium|low",\newline
\ \ \ \textcolor{jsongray}{"timing\_match"}: "good|fair|poor",\newline
\ \ \ \textcolor{jsongray}{"reasoning"}: "[40 words explaining decision]",\newline
\ \ \ \textcolor{jsongray}{"result"}: \textcolor{jsonblue}{"Play or Skip"}\newline
\} \\

\midrule
\textbf{Pattern} & 
Deduces nuanced structural dynamics. Guides the evaluator to isolate systematic implicit signals from interaction history, such as completion rates or re-watching loops. & 
\ttfamily ... Your task is to predict if user play or skip a row based on recent interactions.\newline
...\newline
\#Guidelines to reasoning: \textcolor{promptorange}{\textbf{consider genre, progress indicators, and rewatching tendencies}} ... & 
\ttfamily \{\newline
\ \ \ \textcolor{jsongray}{"reason"}: "your reasoning here",\newline
\ \ \ \textcolor{jsongray}{"result"}: \textcolor{jsonblue}{"Play or Skip"}\newline
\} \\

\bottomrule
\end{tabularx}
\caption{Prompt template formats for LLM recommendation evaluation. Templates range from baseline direct binary classification to structured frameworks incorporating architectural use-cases such as temporal analysis, user persona tracking, evidence-grounded schemas, and explicit pattern extraction.}
\label{tab:prompt-formats}
\end{table*}


While complex, multi-stage reasoning formats provide the distinct operational advantage of producing rich natural-language rationales—serving as valuable side information for debugging and model interpretability—the primary objective of this evaluation remains the baseline predictive accuracy of the LLM judge. Consequently, during the inference phase, models are measured strictly on their downstream capability to accurately predict the final user engagement label under each respective prompting paradigm.

\subsection{Training Configurations}
In RQ2, we aim to understand whether the LLM's predictive accuracy can be improved by training it on a task-specific dataset. We evaluated three training approaches: Supervised Fine-tuning (SFT), Direct Preference Optimization (DPO) \cite{rafailov2023direct}, and a sequential regime consisting of SFT followed by DPO. For each training approach, we compared a simple prompt formulation with a prompt that included reasoning instructions.

\subsubsection{Supervised Fine-Tuning (SFT)}
We fine-tuned the models using both direct prediction (simple prompt) and reasoning-based prediction task formulations. In the direct prediction formulation, training data consisted of examples formatted with a simple prompt (no reasoning instructions) with output targets consisting of just the correct final label. In the reasoning formulation, training data consisted of examples formatted with a reasoning-instruction prompt and output targets consisting of both a ground truth reasoning path and a final label. Ground truth reasoning paths were generated by a larger LLM with reasoning capabilities (see Section \ref{reasoning_datagen}). 

\subsubsection{Direct Preference Optimization (DPO)}
We also experimented with DPO to more explicitly align the model to prefer correct responses over incorrect responses. As above, we tested both direct prediction (simple prompt) and reasoning-based prediction task formulations. In the simple formulation, preference pairs are induced by simply selecting the correct response over the incorrect response (e.g., ``play'' over ``skip'' if the user actually played). In the reasoning formulation, we constructed preference pairs by generating reasoning paths for both potential outcomes, ``play'' or ``skip'' (see Section \ref{reasoning_datagen}). For each training instance, the ``chosen'' response was the generated reasoning and label that matched the actual user action, while the ``rejected'' response was the generated reasoning and label for the opposite action.

\subsubsection{SFT + DPO}
We also tested a sequential training regime consisting of SFT followed by DPO. This regime is a common two-stage recipe used throughout the industry for aligning LLMs with product goals; SFT is used first to establish the structural format and domain vocabulary, then DPO further optimizes decision-making.

\subsection{Generating Reasoning Traces}\label{reasoning_datagen}
To provide the behavioral alignment signal required for fine-tuning or preference optimization, we constructed a synthetic training dataset of paired correct and counterfactual rationales using a high-capacity reasoning LLM as a teacher. Because unaligned LLMs inherently function as bidirectional rationalizers (as characterized in Section \ref{sec:rationalizer}), a single teacher model can reliably generate coherent, factually grounded arguments for both potential outcomes from the exact same user history.

For each instance, we provided the teacher model with the serialized user history, the recommended row, and a candidate engagement label (``play'' or ``skip''). We then prompted the teacher to generate a definitive, step-by-step rationale explaining the candidate outcome. By systematically generating reasoning paths for both potential outcomes, we obtained a pair of rationales for every instance: one matching the true user behavior and a counterfactual one arguing the opposite. During SFT, we trained the model exclusively on the rationales corresponding to the true user actions. For DPO, we leveraged the paired data, designating the rationale aligned with the true action as the ``chosen'' response and the counterfactual rationale as the ``rejected'' response.

\section{Experimental Setup}
\subsection{Data} \label{sec:data}

We conducted our experiments using real-world historical interaction logs from a production recommendation system. Our dataset consists of verified impression events, capturing instances where a user is exposed to curated recommendation rows. To construct a reliable ground truth, we use a spatial heuristic based on how users scroll. When a user navigates a multi-row homepage and plays an item from a specific row, we assume the user saw that row and all the rows positioned above it. Consequently, the row containing the interacted item is logged as a positive ``play'' event; rows that the user scrolled past are logged as negative ``skip'' events. For each impression, our data captures the user's chronological watch history up to the impression event, the immediate session context, the specific items displayed in that single row, and its corresponding ground-truth label.

\subsection{Baselines} \label{sec:eval}

We evaluate our LLM-based judge against two baselines: (1) a zero-shot LLM-based evaluator using the simple prompt format, and (2) a heavily feature-engineered production baseline used internally for offline evaluation. 

For the zero-shot baseline, we use Llama 3.1 8B, which offers a favorable trade-off between capability and efficiency, delivering high-quality predictions while remaining small enough to be considered as a candidate for deployment in production systems. In particular, the baseline zero-shot approach was three times more likely to predict a ``play'' result over a ``skip'' result in our balanced dataset (see Table \ref{tab:prompt_results}). This bias resulted in high recall for play events but low overall precision, and demonstrates a key limitation of unaligned LLM judges for personalized recommendation evaluation: pre-trained LLMs have no notion of the underlying baseline take rates of our particular recommendation task and therefore fail to reflect realistic selection probabilities.  

The production baseline combines an extensive, heavily engineered feature pipeline with a neural network producing a score between 0 and 1, representing the overall quality and relevance of a recommended row for a specific user. Comparison against the production baseline serves two purposes. First, it establishes capability and provides clarity for the industry: can an LLM-based evaluator match or surpass the baseline's accuracy using only raw, serialized text logs? Second, it surfaces qualitative benefits unique to the LLM approach: interpretable reasoning traces that expose the user-history signals driving each prediction, no manual feature pipeline to maintain, and a unified evaluator adapts to new recommendation contexts without re-engineering features.

We report Macro F1-Score as our primary evaluation metric, as it balances recall and precision across both ``Play'' and ``Skip'' classes.

\section{Results} \label{sec:results}

\subsection{Prompting Dynamics (RQ1)}
Through iterative prompt engineering, we identified several key dynamics regarding input design. 
First, we found that session context, such as the current time, proved to be among the most critical factors to include in the prompt. This finding is consistent with the literature on traditional, non–LLM-based recommender systems, which emphasizes the importance of context awareness for effective recommendation. In our setting, the LLM is able to infer and exploit the relevant context directly from the prompt and appears to rely strongly on this immediate context when predicting engagement.

\begin{table}[t]
  \caption{Macro-F1 lift (\%) of each prompt format over the simple prompt baseline (higher is better) and Positive Bias (ratio of ``Play'' to ``Skip'' predictions; 1.0 indicates no bias). Results are evaluated in a zero-shot setting using Llama 3.1 8B. See Table \ref{tab:prompt-formats} for details on prompt formulations.}
  \label{tab:prompt_results}
  \centering
     \resizebox{\linewidth}{!}{%
  \begin{tabular}{lcc}
    \toprule
    \textbf{Format} & \textbf{Macro-F1 Lift (\%)} & \textbf{Positive Bias ($\times$)} \\
    \midrule
    Simple & - & 3.01 \\ 
    Simple\textsuperscript{-1} & -26.79 & 20.08\\
    Reason & \bf{4.21} & 0.79 \\
    Reason\textsuperscript{-1} & -7.39 & 0.34 \\
    Evidence & -25.89 & 24.20 \\
    Pros \& Cons & 0.76 & 0.90 \\
    Habit & -28.83 & 34.23 \\
    Persona & -18.40 & 5.76 \\
    Timing & -13.75 & 7.55 \\
    Pattern & -12.33 & 5.14 \\
    \bottomrule
  \end{tabular}
  }
\end{table}

\begin{table*}[t]
  \caption{Macro-F1 lift (\%) over the Llama 3.1 8B zero-shot baseline (higher is better) and Positive Bias (ratio of ``Play'' to ``Skip'' predictions; 1.0 indicates no bias) for each training and inference configuration. The best Macro-F1 lift (bolded) is SFT+DPO with reasoning-based inference, at 32.19\%.}
  \label{tab:results_and_bias}
  \centering
  \begin{tabular}{llcccc}
    \toprule
      & & \multicolumn{2}{c}{\textbf{Macro-F1 lift (\%)}} & \multicolumn{2}{c}{\textbf{Positive Bias ($\times$)}} \\
      \cmidrule(lr){3-4}\cmidrule(lr){5-6}
    &  & \multicolumn{2}{c}{\textbf{Inference prompt}} & \multicolumn{2}{c}{\textbf{Inference prompt}} \\
    \textbf{Training paradigm} & \textbf{Training prompt} & \textbf{Simple} & \textbf{Reason} & \textbf{Simple} & \textbf{Reason} \\
    \midrule
    Zero-shot   & -      & 0.0    & 4.21        & 3.01  & 0.79 \\
    SFT         & Simple & 12.23  & 11.09       & 1.79  & 1.51 \\
    SFT         & Reason & -15.84 & 11.35       & 10.66 & 1.61 \\
    DPO         & Simple & 28.37  & 20.40       & 0.70  & 0.54 \\
    DPO         & Reason & 23.35  & 23.56       & 1.18  & 1.18 \\
    SFT + DPO   & Simple & 30.58  & 19.41       & 0.78  & 0.89 \\
    SFT + DPO   & Reason & 25.74  & \bf{32.19}  & 0.93  & 1.05 \\
    \bottomrule
  \end{tabular}
\end{table*}

To understand whether positive class bias can be mitigated via prompt engineering, we tested an inverse version of the prompt, which presented the task as predicting whether a user will skip, rather than play, the recommended row. Surprisingly, this change further \textit{increased} the positive class bias, and led to a large degradation in overall performance (see Table \ref{tab:prompt_results}). One hypothesis is that changing the task to predict whether the user will skip \textit{all} recommended items in a row becomes more difficult than the original task of predicting whether the user will play \textit{any} recommended item in a row. 

We also evaluated the impact of context length by varying the number of events included in the user history. Using a longer user history was associated with performance gains up to a certain point, after which performance plateaus. Adding more historical events beyond this threshold did not help the model, indicating a point of diminishing returns where the LLM struggles to effectively utilize overly long historical contexts. For the remaining experiments, we limited user history to 50 events.

Finally, we found that prompting the model to reason before outputting a final label yielded a noticeable improvement in predictive performance compared to direct label prediction. This increased Macro-F1 score 4.21\% over the baseline non-reasoning prompt in the zero-shot setting. Interestingly, the more structured reasoning instructions we tested (see Section \ref{sec:reasoning_prompts}) did not further improve performance over the simple reasoning prompt and in many cases severely degraded performance (see Table \ref{tab:prompt_results}). 
For this reason, in the remaining experiments, we compare only the simple direct prompt with the basic reasoning prompt.

While these prompt engineering choices improve accuracy over the zero-shot baseline, the best-performing prompted configuration still leaves a significant gap relative to the feature-engineered baseline, motivating the move to parameter-level adaptation.

\subsection{Training Paradigms (RQ2)}
Supervised fine-tuning provided a significant boost to predictive performance, increasing Macro-F1 Score by 12.23\% over the zero-shot baseline in the simple prompt setup. We hypothesize that the majority of this improvement comes from the LLM learning to more effectively calibrate baseline engagement probabilities, reducing the over-prediction of ``play'' events---after SFT the positive class bias reduces from 3x over-prediction of ``play'' events to only 1.79x (see Table \ref{tab:results_and_bias}). 
While the reasoning-prompt SFT paradigm also improved performance over the zero-shot baseline (+11.35\%), we did not see any additional gains by training with a reasoning prompt and including synthetically generated reasoning paths in the training data when compared to the simple-prompt SFT paradigm. In fact, when prompted again with the simple non-reasoning prompt the model fine-tuned on reasoning traces performed substantially worse than zero-shot (-15.84\%), showing signs of task-specific overfitting. In contrast, the model fine-tuned with the simple prompt set-up was still able to outperform the zero-shot reasoning prompt performance when prompted with a reasoning instruction at inference time (+11.09\%).

Training the model with DPO proved more effective than SFT alone, leading to an improvement of 28.37\% in Macro-F1 score over the zero-shot baseline with the simple prompt setup (i.e., trained to prefer the correct ``play'' or ``skip'' response over the incorrect one). As was the case with SFT, the reasoning prompt training paradigm with DPO improved performance over the zero-shot baseline (+23.56\%), but provided no additional gains over the simple prompt training paradigm.

We achieved the best overall results in our experiments by sequentially chaining the two methods: first utilizing SFT to establish the domain vocabulary and formatting, followed by DPO to optimize the decision-making process based on accurate versus inaccurate reasoning. The SFT+DPO combination with reasoning-based prediction yields a 32.19\% Macro-F1 lift over the zero-shot baseline---our best result, closing the gap to the feature-engineered production baseline by reaching statistical parity in Macro-F1 score (difference < 0.1\%).

\section{Conclusion} \label{sec:conclusion}

In this work, we characterized \emph{bidirectional rationalization} as a personalization-specific failure mode of LLM-based evaluators that is structurally distinct from hallucination and rooted in fundamental recommender-system trade-offs. Because these trade-offs admit multiple defensible interpretations of the same user evidence, an unaligned judge can produce coherent rationales for opposing predictions. 

We further show that prompt engineering is insufficient to eliminate this failure mode. While prompt design can improve over a zero-shot LLM judge, it remains well below a heavily feature-engineered baseline. To close this gap, we use an alignment recipe that combines reasoning-based prediction with preference optimization over paired correct and counterfactual rationales anchored to observed engagement outcomes. The resulting text-based LLM evaluator matches the feature-engineered baseline on Netflix homepage engagement prediction without manual feature engineering and produces human-interpretable reasoning traces that expose the user-history signals driving each predicted outcome. 

Overall, our results suggest that personalized LLM judges fail not because they fabricate evidence, but because they apply underspecified, competing framings of identical evidence. Behavioral alignment via paired-rationale preference is key to making such judges reliable. 

Future work will explore intermediate user profile generation to address the diminishing returns observed when extending raw user histories beyond 50 events. Instead of directly introducing extensive, raw interaction logs into the model's context window, condensing long-term user tastes and behavioral patterns into concise, natural-language profiles could allow the evaluator to leverage much deeper historical signals without suffering from attention dilution.


\bibliographystyle{ACM-Reference-Format}
\bibliography{custom}



\end{document}